\documentclass[lettersize,journal]{IEEEtran}
\usepackage{amsmath,amsfonts}
\usepackage{algorithmic}
\usepackage{algorithm}
\usepackage{array}
\usepackage[caption=false,font=normalsize,labelfont=sf,textfont=sf]{subfig}
\usepackage{textcomp}
\usepackage{stfloats}
\usepackage{url}
\usepackage{verbatim}
\usepackage{graphicx}
\usepackage{dsfont}
\usepackage{cite}
\usepackage{color}
\usepackage{ulem}
\usepackage{booktabs}
\usepackage{threeparttable}
\usepackage{utfsym}
\usepackage{bbding}
\usepackage[switch]{lineno}

\usepackage{tikz,xcolor,hyperref}
\definecolor{lime}{HTML}{A6CE39}
\DeclareRobustCommand{\orcidicon}{%
    \begin{tikzpicture}
    \draw[lime, fill=lime] (0,0) 
    circle [radius=0.16] 
    node[white] {{\fontfamily{qag}\selectfont \tiny ID}};    \draw[white, fill=white] (-0.0625,0.095) 
    circle [radius=0.007];    \end{tikzpicture}
    \hspace{-2mm}}
\foreach \x in {A, ..., Z}{%
    \expandafter\xdef\csname orcid\x\endcsname{\noexpand\href{https://orcid.org/\csname orcidauthor\x\endcsname}{\noexpand\orcidicon}}
    }

\newcommand{\wh}[1]{\textcolor{black}{{#1}}} 

\newcommand{\khb}[1]{\textcolor{black}{{#1}}}
\newcommand{\kkhb}[1]{\textcolor{black}{{#1}}}
\newcommand{\yl}[1]{{\textcolor{black}{#1}}} 

\begin{document}

\title{DyCrowd:  Towards Dynamic Crowd Reconstruction from a Large-scene Video}

\author{Hao Wen$^{\dagger}$\orcidB{}, 
Hongbo Kang$^{\dagger}$\orcidA{}, 
Jian Ma\orcidC{}, 
Jing Huang\orcidD{}, 
Yuanwang Yang\orcidE{}, 
Haozhe Lin\orcidH{}, \\
Yu-Kun Lai\orcidF{} {~\IEEEmembership{Senior Member,~IEEE}},
Kun Li$^{*}$\orcidG{} {~\IEEEmembership{Senior Member,~IEEE}}
\thanks{$\dagger$ Equal contribution.}
\thanks{$*$ Corresponding author.}
\thanks{Hao Wen, Hongbo Kang, Jian Ma, Jing Huang, Yuanwang Yang and Kun Li are with the College of Intelligence and Computing, Tianjin University, Tianjin 300350, China. E-mail: \{wenhao, hbkang, jianma, hj00, yyw, lik\}@tju.edu.cn}
\thanks{Haozhe Lin is with Beijing National Research Center for Information Science and Technology, Tsinghua University, Beijing 10084, China. E-mail: linhz@tsinghua.edu.cn}
\thanks{Yu-Kun Lai is with the School of Computer Science and Informatics, Cardiff University, Cardiff CF24 4AG, United Kingdom. E-mail: Yukun.Lai@cs.cardiff.ac.uk}
}

\markboth{IEEE TRANSACTIONS ON PATTERN ANALYSIS AND MACHINE INTELLIGENCE}%
{Shell \MakeLowercase{\textit{et al.}}: A Sample Article Using IEEEtran.cls for IEEE Journals}


\maketitle
\begin{figure*}[t!]
\centering
\includegraphics[width=\textwidth]{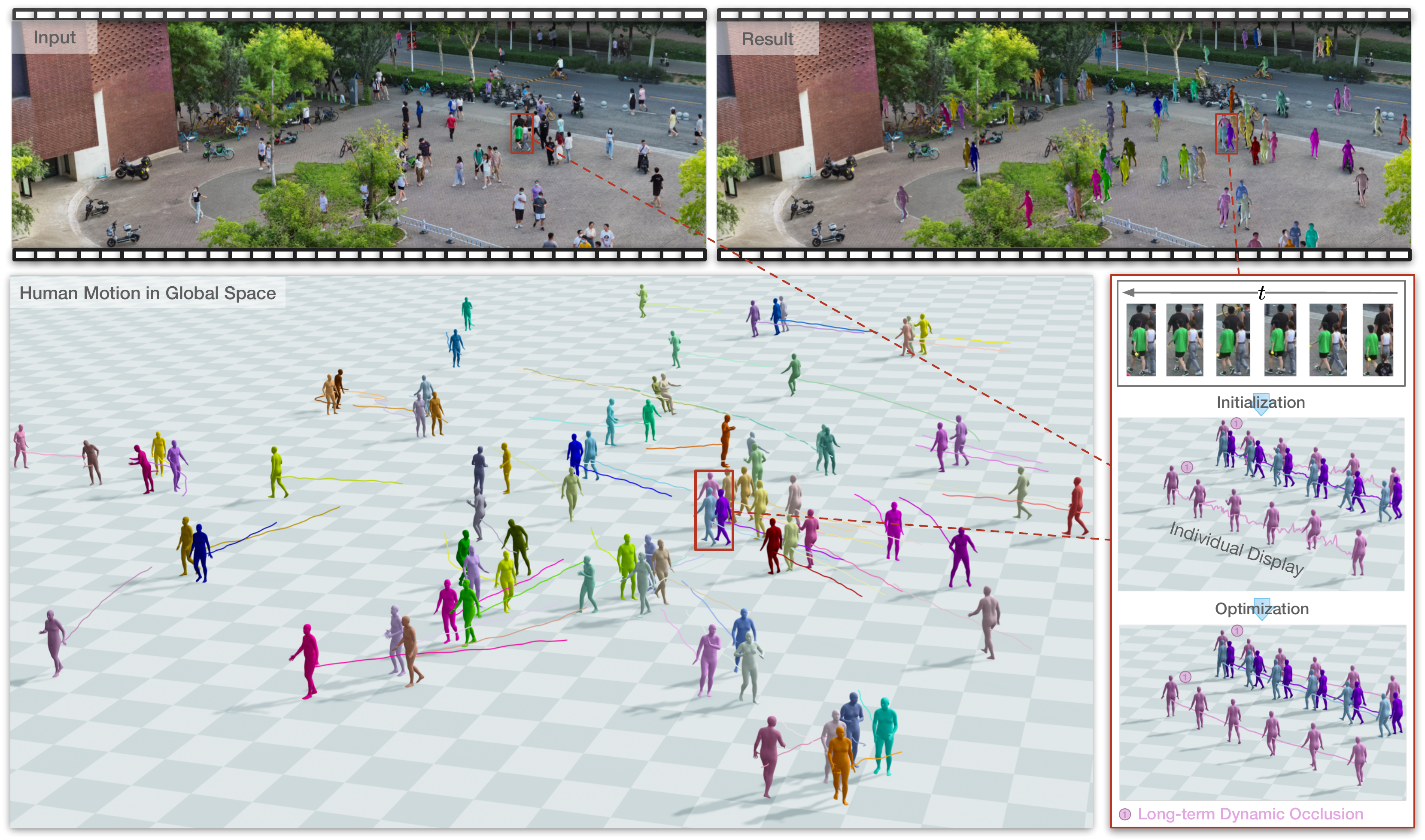}
\caption{Spatio-temporally consistent 3D crowd reconstruction from a large-scene video via the proposed DyCrowd. This approach effectively achieves dynamic 3D reconstruction in a global space and addresses frequent dynamic occlusions in large scenes. Even when certain subjects (e.g., subject 1 \yl{in the figure}) remain segment-level occluded throughout the entire video, our method can still reconstruct reasonable motion trajectories and poses through optimization.}
\label{fig:fig1}
\end{figure*}

\begin{abstract}
3D reconstruction of dynamic crowds in large scenes has become increasingly important for applications such as city surveillance and crowd analysis. However, current works attempt to reconstruct 3D crowds from a static image, causing a lack of temporal consistency and  \yl{inability} to alleviate the typical impact caused by occlusions. In this paper, we propose DyCrowd, the first framework for spatio-temporally consistent 3D reconstruction of hundreds of individuals' poses, positions and shapes from a large-scene video. We design a coarse-to-fine group-guided motion optimization strategy for occlusion-robust crowd reconstruction in large scenes. \yl{To address temporal instability and severe occlusions,}
\yl{we further incorporate} a VAE \yl{(Variational Autoencoder)}-based human motion prior along with a segment-level group-guided optimization. 
The core of our strategy leverages collective crowd \yl{behavior} to address long-term dynamic occlusions. By jointly optimizing the motion sequences of individuals with similar motion segments and combining this with the proposed Asynchronous Motion Consistency (AMC) loss, we enable high-quality unoccluded motion segments to guide the motion recovery of occluded ones, ensuring robust and plausible motion recovery even in the presence of temporal desynchronization and rhythmic inconsistencies.
Additionally, in order to fill the gap of no \yl{existing well-annotated} large-scene video dataset, we contribute a virtual benchmark dataset, \textit{VirtualCrowd}, for evaluating dynamic crowd reconstruction from large-scene videos. Experimental results demonstrate that the proposed method achieves state-of-the-art performance in the large-scene dynamic crowd reconstruction task. The code and dataset will be \yl{available} for research purposes.
\end{abstract}


\section{Introduction}
\IEEEPARstart{3}{D} human pose and shape estimation in large scenes is essential for enabling effective \yl{behavioral modeling} and crowd monitoring. 
This capability plays a pivotal role in a wide range of applications, from public safety and urban planning to large-scale event management. However, current efforts~\cite{wen2023crowd3d, huang2023reconstructing} related to large-scene 3D pedestrians reconstruction are based on a single image, capturing static crowds' 3D positions, postures and shapes.
While these methods can provide useful insights into the spatial distribution and individual characteristics of crowds, they fall short when it comes to representing temporal dynamics and continuous movement patterns, which are critical for understanding crowd \yl{behavior} over time.
This paper aims to reconstruct dynamic crowds within a global camera space from a large-scene video.

Most single-person reconstruction methods estimate temporally consistent and smooth pose and shape by using various temporal modeling techniques~\cite{kocabas2020vibe, choi2021beyond, wei2022capturing_mpsnet, shen2023global} or optimization paradigms~\cite{yuan2022glamr, wang2024tram} from a single RGB video.
To extend to large-scene crowd videos, an intuitive idea is to combine these methods with multi-object tracking techniques and independently reconstruct multiple people from cropped instance images~\cite{wen2023crowd3d,li2023mili}. 
However, \yl{this approach} ignores interactions among individuals and overlooks critical location and scale information within the global coordinate system of large scenes.
For video-based multi-person reconstruction works, the regression-based \yl{approaches}~\cite{sun2023trace, qiu2023psvt} reconstruct multiple people at once from a video, \yl{but they fail} to cope with large scenes because scaling the high-resolution images of large scenes down to input resolution would result in the loss of image features for most small and medium-sized individuals. Moreover, insufficient data availability hinders regression-based methods from effectively reconstructing crowds from large-scene videos. 
Additionally, existing optimization-based methods~\cite{ye2023decoupling, kocabas2024pace}, designed specifically for small scenes with handheld cameras, cannot be directly applied to large-scene videos due to unique challenges such as a large number of people, severe scale variation, wide spatial distribution, and frequent dynamic occlusions.

Crowd3D~\cite{wen2023crowd3d} and GroupRec~\cite{huang2023reconstructing} are two state-of-the-art methods for multi-person reconstruction in large scenes, predicting human body meshes with absolute positions for hundreds of pedestrians within a unified camera system from a single large-scene image. However, when applied to each frame of a large-scene video, these methods often yield temporally-unstable and \yl{unrealistic} 3D motion and frequently lose objects due to severe mutual occlusion.
Occlusions frequently occur and may last for a long time in large scene videos due to the high density of moving individuals, causing spatio-temporal discontinuity and thereby hampering dynamic human reconstruction in large scenes. 

In this paper, we propose DyCrowd, the first framework capable of reconstructing the 3D positions, poses, and shapes of dynamic crowds from large-scene videos. To achieve robust and plausible crowd reconstruction in large scenes, we design a Coarse-to-fine Group-guided Motion (C.G.M.) optimization strategy, which includes VAE-based human motion prior optimization and segment-level group-guided optimization. This strategy fully leverages the collective behavior of crowds in large scenes~\cite{zhou2013measuring, wu2017collective, mei2019measuring}: (1) pedestrians with similar trajectories often exhibit similar motion postures, and (2) severely occluded pedestrians and unoccluded pedestrians frequently coexist in the same scene. 
\khb{To effectively leverage the collective motion within groups, unlike GroupRec~\cite{huang2023reconstructing}, which reconstructs crowds through collective behavior in static images, we focus on the dynamic changes of the group across consecutive frames and the commonalities in motion among individuals. The goal is to utilize high-quality and unoccluded motion data to assist in recovering the motion of occluded pedestrians.} Therefore, we first cluster individual motion segments based on trajectory similarity and introduce detection scores and joint importance to determine the confidence of motion segments, identifying those requiring correction and high-quality, unoccluded motion segments. It is important to note that a segment of an individual's motion may be mixed with other segments from the same or different individuals in the same group. To address issues such as temporal misalignment and rhythm variations among motion sequences within the group, we propose an Asynchronous Motion Consistency (AMC) loss, ensuring the successful alignment and recovery of long-term occluded motion sequences under the guidance of unoccluded motion. To ensure the quality of group motion segments in C.G.M., we employed a transformer Variational Autoencoder (VAE) and trained the model using occlusion data augmentation to develop a human motion prior, thereby improving the smoothness and authenticity of the motion.

Furthermore, to verify the proposed method, we create a synthetic large-scale dynamic crowd dataset \textit{VirtualCrowd} by employing motion generation techniques~\cite{zhang2022wanderings} and datasets of virtual humans~\cite{zhao2023synthesizing} and scenes~\cite{icity3d}. 
\khb{This dataset provides eight 8K validation videos from four large-scale scenes, featuring varying crowd densities and perspectives (high-angle and low-angle), with detailed 2D/3D annotations, including 2D and 3D keypoints, global positions, and SMPL-X~\cite{pavlakos2019expressive} parameters.}
\textit{VirtualCrowd} is expected to facilitate future research on large-scene dynamic crowd reconstruction and analysis.
Experimental results on both synthetic and real captured videos demonstrate that our method achieves spatio-temporally consistent 3D crowd reconstruction from a large-scene video, as shown in Fig. \ref{fig:fig1} with an example. The code and dataset \yl{will be}
publicly available for research purposes.

In summary, the main contributions of this paper are:
\begin{itemize} 
    \item This is the first framework for reconstructing the motion of crowds from a large-scene video, achieving spatial and temporal consistency, and continuous and natural reconstruction of 3D positions, poses and shapes of hundreds of people.
    \item We propose a coarse-to-fine group-guided motion optimization strategy to fit crowd motion, by introducing a VAE-based motion prior in optimization followed by a  segment-level group-guided optimization.
    \item We design an asynchronous motion consistency loss that measures the differences in motion segments within a group. This helps in recovering motion under long-term occlusions.
    \item We contribute a virtual dataset that includes four dynamic scenes with an average of hundreds of people, equipped with detailed 3D labels such as 3D joints, 3D positions, and SMPL-X parameters to evaluate large-scene dynamic crowd reconstruction.
\end{itemize}

\section{RELATED WORK}

\subsection{Crowd Reconstruction in Large Scenes}
In recent years, the field of imaging has made significant advancements, producing gigapixel-level images with a wide field of view and high resolution details. These images are typically captured by a multi-scale camera array ~\cite{brady2012multiscale, yuan2017multiscale}.
A notable dataset in this domain is PANDA~\cite{wang2020panda} which is a human-centric large-scene video dataset, including high-resolution ($\sim$25,000$\times$15,000) images of large-scale scenes with hundreds of people, and has been applied across a spectrum of visual tasks~\cite{liu2024gigahumandet,mo2023pvdet,li2024saccadedet,zhang2023towards,li2023mili}. 

Particularly, building on the availability of large-scale scene data, 
Wen et al.\cite{wen2023crowd3d} introduce the task of crowd reconstruction in a large scene, which aims to reconstruct 3D poses, shapes, and positions of hundreds of individuals within a unified camera space from a single large-scene image.
They design a joint local and global inference framework called Crowd3D,  which converts the complex crowd localization into pixel localization with the help of Human-scene Virtual Interaction Point (HVIP) and the parameters of pre-estimated scene-level camera and ground plane, directly reconstructing human body meshes in a large-scene camera system from cropped images.
Moreover, GroupRec~\cite{huang2023reconstructing} solves the task by incorporating group features and positional information into network inference, accounting for collective \yl{behaviors} and interactions to derive relatively accurate crowd positioning. 
However, these methods solely focus on a single-image reconstruction and as a result, they are unable to infer natural and smooth crowd motion from a large-scene video.

\subsection{Video-based Single-person Reconstruction}

Typically, reconstructing a single person from a video is a reasonable way to maintain the consistency of the pose and shape over time. Previous \yl{research}~\cite{kocabas2020vibe, choi2021beyond, wei2022capturing, xu20213d, shen2023global} primarily \yl{focuses} on exploring various temporal  modeling architectures.
VIBE~\cite{kocabas2020vibe} and TCMR~\cite{choi2021beyond} rely on recurrent neural \yl{networks (RNNs)} for temporal information. VIBE builds a bidirectional gated recurrent unit (GRU) time encoder and enhances the plausibility of motion sequences through adversarial training. However, the reconstruction results are more heavily dependent on static features.
To address it, TCMR aggregates temporal features of the past, future, and \yl{current} frame 
to strongly constrain outputs with motion consistency.
Compared to using an RNN, MPS-Net~\cite{wei2022capturing} captures motion continuity dependencies through attention mechanisms, enhancing temporal correlations.
Due to the inability of attention mechanisms to effectively capture local details, GLoT~\cite{shen2023global} utilizes transformer architectures to encode videos, capturing long-term temporal correlations while refining local details through attention on nearby frames.
Recent approaches\cite{yuan2022glamr, shin2024wham, wang2024tram} focus on estimating the global human trajectory from a monocular dynamic camera.
GLAMR~\cite{yuan2022glamr} utilizes a transformer-based generative motion filler to fill missing motion from regression.
It optimizes global trajectories between multiple frames of individuals in the scene but ignores the camera motion cues.
Both WHAM~\cite{shin2024wham} and TRAM~\cite{wang2024tram} recover camera motion with SLAM (Simultaneous Localization and Mapping). 
WHAM recursively predicts the pose, shape, and global motion parameters without optimization for faster inference. However, it may have poor generalization for novel posture as it learns human motion from MoCap data.
TRAM effectively captures more accurate human motion by composing camera trajectory in the world frame and a person's body motion in the camera frame, \yl{achieving} better generalization to complex scenarios.

However, these methods struggle with large-scene videos featuring hundreds of people because they estimate the pose and shape of each person in local camera space, resulting in an incoherent spatial distribution of the reconstructed multi-person meshes. Additionally, they neglect interactions between multiple individuals and struggle to handle scenarios involving mutual occlusion.

\subsection{Video-based Multi-person Reconstruction}

Unlike methods that reconstruct humans on a person-by-person basis, some approaches focus on recovering multi-person meshes simultaneously. These methods demonstrate the ability to reconstruct globally consistent positions and scale relationships among individuals, offering greater robustness in handling mutual occlusions and preserving spatial coherence across the scene.
\yl{Benefiting} from the exploration of one-stage multi-person pose and shape estimation~\cite{zhang2021body, sun2021monocular, sun2022putting}, some methods~\cite{qiu2023psvt, sun2023trace} have achieved end-to-end prediction of multiple people in a video.
Specifically, PSVT~\cite{qiu2023psvt} employs a progressive video Transformer encoder to capture global feature dependencies among 
\yl{subjects},
enabling end-to-end estimation of multiple \yl{people's} poses and shapes. 
TRACE~\cite{sun2023trace} utilizes optical flow as a motion cue and estimates multiple people by a novel 5D representation that represents the temporal motion of a person in both camera and global coordinates.
Regrettably, these methods are unable to directly regress humans from an entire large-scene image because, compared to the image size, the scale of humans is relatively small and varies greatly.

Other methods~\cite{kocabas2024pace, ye2023decoupling} utilize multi-stage spatio-temporal joint optimization, incorporating generative models~\cite{rempe2021humor, he2022nemf} as motion priors, to reconstruct spatially and temporally consistent multi-person poses and shapes from a video.
PACE~\cite{kocabas2024pace} integrates SLAM and human motion priors into an optimization framework, leveraging the strengths of both to jointly optimize the motions of humans and the camera.
SLAHMR~\cite{ye2023decoupling} decouples camera and human motion by integrating relative camera estimation and learned transition human motion priors to address the scene scale ambiguity, thereby inferring the motion of multiple individuals in a shared coordinate system. Although multi-stage optimization processes may appear suitable for large-scene videos, these methods are typically tailored for small scenes captured by hand-held cameras and are not equipped to address the unique challenges of large scenes, such as significant scale variations, complex spatial extents, and frequent mutual occlusions. 

In this paper, we introduce DyCrowd, the first framework for reconstructing spatio-temporally consistent 3D positions, poses, and shapes of hundreds of people in a unified global space from a large-scene video. 
DyCrowd produces natural crowd motion and addresses the problem of motion recovery under long-term occlusion in large-scene videos through our coarse-to-fine group-guided motion optimization strategy and asynchronous motion consistency loss.
We also contribute a virtual dataset called \textit{VirtualCrowd} to support the evaluation of large-scene dynamic crowd reconstruction.

\section{Methods}

\begin{figure*}[t!]
\centering
\includegraphics[width=\textwidth]{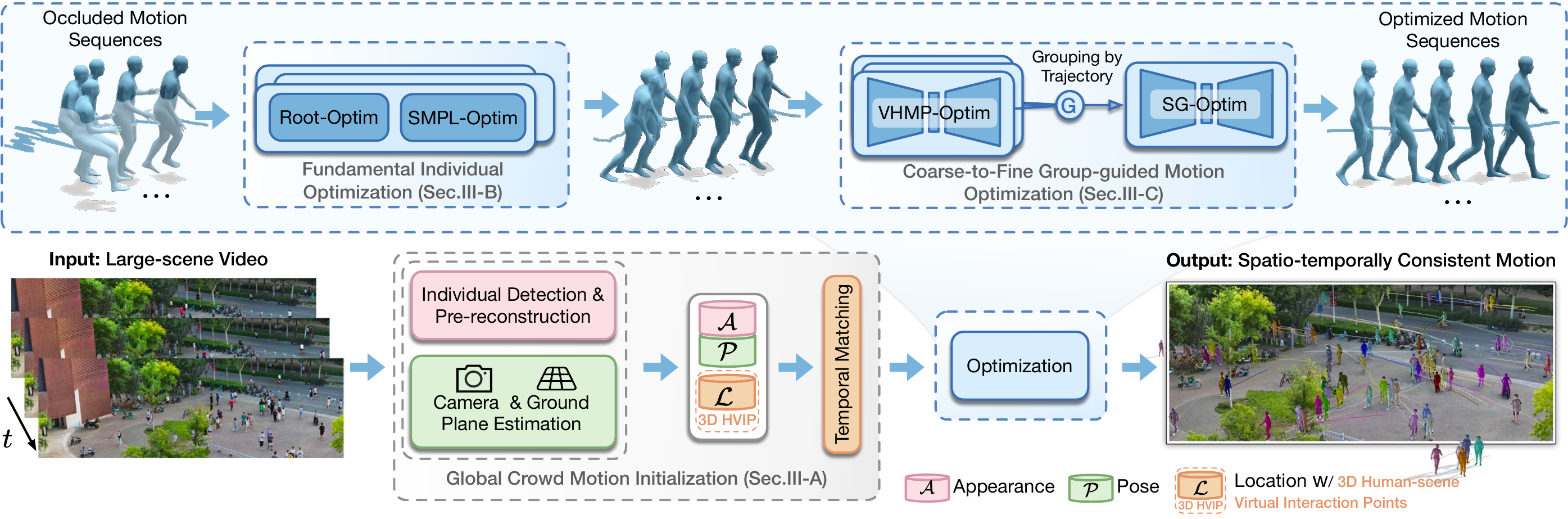}
\caption{Overview of DyCrowd framework. 
DyCrowd is a multi-stage optimization method that consists of three main components: global crowd motion initialization, fundamental individual optimization, and coarse-to-fine group-guided motion optimization. The coarse-to-fine group-guided motion optimization utilizes a VAE-based Human Motion Prior Optimization (VHMP-Optim) to enhance the temporal stability of crowd motion and employs a Segment-level Group-guided Optimization (SG-Optim) to address long-term occlusion issues in large-scene videos. This process results in a spatio-temporally consistent 3D reconstruction of crowd motion within a unified global space.}
\label{fig:fig2}
\end{figure*}

The proposed DyCrowd aims to achieve precise and continuous 3D reconstruction of the global positions, poses, and shapes of hundreds of people from a large-scene video. Fig.~\ref{fig:fig2} illustrates the framework of our proposed approach. The core of our approach is a coarse-to-fine group-guided motion optimization strategy that addresses temporal instability through a human motion prior model and restores long-term occluded postures by leveraging the collective \yl{behavior} of the crowd.
DyCrowd employs a multi-stage optimization process to achieve spatial-temporal consistent global crowd dynamics reconstruction:
1) integrating estimations from data-driven models to obtain the global crowd motion initialization (Sec.~\ref{sec:global-init});
2) employing a fundamental individual optimization for reliable positions, poses and shapes of individuals (Sec.~\ref{sec:fundamental});
3) addressing motion instability and dynamic occlusions by our coarse-to-fine group-guided motion optimization strategy (Sec.~\ref{sec:group-opt}), which progressively optimizes natural and occlusion-robust dynamic crowd motion.

We take as input a large-scene video with $T$ frames of a scene with
$N$ individuals and represent their motion sequence in a global large-scene coordinate system with $\{\mathbf{Q}_n\}_{n=1}^N$.
We use SMPL~\cite{SMPL:2015} model and $\mathbf{Q}_n = \{{Q}_{n,t}=\{\boldsymbol{\theta}_{n,t}, \boldsymbol{\beta}_{n,t},\boldsymbol{\gamma}_{n,t}, \boldsymbol{\tau}_{n,t} \}\}_{t=1}^{T}$.
For each individual, the SMPL model transforms body pose parameters $\boldsymbol{\theta}_{n,t} \in \mathbb{R}^{23 \times 3}$, shape parameters $\boldsymbol{\beta}_{n,t} \in \mathbb{R}^{10}$, rotation parameters $\boldsymbol{\gamma}_{n,t} \in \mathbb{R}^{3}$ and translation $\boldsymbol{\tau}_{n,t} \in \mathbb{R}^{3}$ into the human body mesh vertices $\mathbf{V}_{n,t} \in \mathbb{R}^{6890 \times 3}$ and joint positions $\mathbf{J}_{n,t}^{\text{SMPL}} \in \mathbb{R}^{24 \times 3}$ by applying a differentiable function $\mathcal{M}$:
\begin{equation}
\label{deqn_ex1a}
[\mathbf{V}_{n,t}, \mathbf{J}_{n,t}^{\text{SMPL}}] = \mathcal{M}(\boldsymbol{\gamma}_{n,t}, \boldsymbol{\theta}_{n,t}, \boldsymbol{\beta}_{n,t}) + \boldsymbol{\tau}_{n,t}.
\end{equation}

\subsection{Global Crowd Motion Initialization}
\label{sec:global-init}

In large-scene crowd motion reconstruction, we begin by obtaining an initial estimate of the global crowd motion to facilitate subsequent optimization.
This process involves individual detection and pre-reconstruction, camera and ground plane estimation, and temporal matching of individuals to generate preliminary global motion results, which serve as inputs for subsequent optimization.

In the process of individual detection and pre-reconstruction, extracting individual features presents significant challenges due to variations in pedestrian scale caused by differences in their distance from the camera, especially when individuals appear small.
To address this issue, we first employ an adaptive human-centric cropping method~\cite{wen2023crowd3d}, which divides each video frame into multiple regions. Within each region, we apply the top-down detection method, VitDet~\cite{li2022exploring}, to identify individuals. The detection results from all regions are then merged to accurately obtain the bounding boxes and masks of each individual. Based on these predicted bounding boxes, we extract the 2D keypoints $\bar{\mathbf{J}}_{n,t}^\text{2D} \in \mathbb{R}^{17 \times 2}$ for each individual using DWPose\cite{yang2023effective}. Additionally, we use the human mesh reconstruction method, HMR2.0\cite{goel2023humans}, to obtain the locally reconstructed SMPL model's shape parameters $\bar{\boldsymbol{\beta}}_{n,t}\in \mathbb{R}^{10}$ and pose parameters $\bar{\boldsymbol{\theta}}_{n,t}\in \mathbb{R}^{23 \times 3}$ for each individual. 

To estimate the ground plane $\bar{\mathbf{G}}:\{\bar{N}_g,\bar{D}_g\}$, and the camera's intrinsic parameters $\bar{\mathcal{K}}$, similar to Crowd3D~\cite{wen2023crowd3d}, we use walking and standing human poses as priors for calibration. Here, $\bar{N}_g$ represents the ground normal, and $\bar{D}_g$ is a constant. Furthermore, we also infer the 2D and 3D Human-scene Virtual Interaction Points (HVIP) \cite{wen2023crowd3d} for each individual by a single-person HVIP regression network and estimated camera parameters, \yl{which are denoted} as: $\bar{\mathbf{p}}_{n,t}^\text{hvip2D} \in \mathbb{R}^{2}$ and $\bar{\mathbf{p}}_{n,t}^\text{hvip3D} \in \mathbb{R}^{3}$. 
HVIP represents the projection point of a person’s 3D torso center on the ground plane, and we use it to locate the person's initial position in the global space.

We employ the PHALP \cite{rajasegaran2022tracking} to achieve inter-frame individual matching.  Unlike \yl{the original} PHALP, we employ 3D HVIP as the position representation to overcome the limitations of the PHALP framework in large-scale scene tracking.
Specifically, PHALP matches individuals frame by frame based on appearance ($\mathcal{A}$), pose ($\mathcal{P}$), and position information ($\mathcal{L}$), and compares these with the current frame's detection results to update trajectories, enabling continuous tracking. However, due to the significant scale differences between small and large scenes, the tracking performance of PHALP in large-scenes is suboptimal, and the position information provided by the framework underperforms in camera setups with a global perspective.
In contrast, 3D HVIP accurately locates individuals on the ground, offering a more effective positional feature. Furthermore, we confine the matching process to spatially adjacent individuals based on the 3D HVIP positional representation, enhancing computational efficiency.

After the global crowd motion initialization, the motion sequence $\bar{\mathbf{Q}}_n$ for each individual is obtained and used as the initial input for the optimization process.

\subsection{Fundamental Individual Optimization}
\label{sec:fundamental}


As $\bar{\mathbf{Q}}_n$ is obtained from local reconstruction inference, its global position, including both rotation and translation, along with the pose of the SMPL model, remain ambiguous. 
To address this, the fundamental individual optimization process focuses on root optimization and SMPL optimization for each frame to ensure robustness and rationality in both position and pose.
Below, we provide a detailed explanation of the optimization formulas and their corresponding objective functions.

The primary objective of root optimization is to obtain a reasonable positional representation by fitting the root joint of the human body. The first set of variables we optimize includes the global root translation $\boldsymbol{\tau}_{n,t}$ and root rotation 
$\boldsymbol{\gamma}_{n,t}$ for each individual. The objective function for this phase is as follows:
\begin{gather}
  \label{eq:root_prior}
\begin{aligned}&\begin{array}{c}\arg\min\\\boldsymbol{\tau}_{n,t}, \boldsymbol{\gamma}_{n,t} : \forall n,t\end{array}E_{\text{root}}, \quad \text{with}\\
 E_{\text{root}} & = E_{\text{2D}} +  E_{\text{hvip2D}} + E_{\text{contact}} + E_{\text{t-trans}}.
\end{aligned}
\end{gather}
The error term $E_{\text{2D}}$ is used to ensure the reconstructed human motion is spatially reasonable and consistent with the image evidence. 
$E_{\text{2D}}$ is measured by the 2D reprojection error between the 3D motion estimation and the 2D body joint observations:
\begin{gather}
\label{eq:project_2d}
E_{\text{2D}}=\lambda_{\text{2D}} \sum_{n,t,j}\sigma_{n,t,j} \delta_{n,t} \left\|\Pi_\mathcal{K} (\tilde{\mathbf{J}}_{n,t,j}^\text{3D})-\bar{\mathbf{J}}_{n,t,j}^\text{2D}\right\|_2^2,
\end{gather}
where $\lambda_\text{2D}$ represents the contribution weight of this term, and $\lambda\yl{_*}$ for the other terms mentioned later also indicates their respective contribution weights.
$\sigma_{n,t,j} \in \{0,1\}$ indicates whether the $j$-th keypoint of the $n$-th individual in the $t$-th frame is valid, $\tilde{\mathbf{J}}_{n,t,j}^\text{3D} \in \tilde{\mathbf{J}}_{n,t}^\text{SMPL}$ is the 3D keypoint \yl{coordinates} to be optimized, $\delta_{n,t}$ is the inverse of scale (bounding box size) of the $n$-th individual in the $t$-th frame, and $\Pi_\mathcal{K}$ is the perspective projection with the camera intrinsic matrix $\mathcal{K}$. To address the limitation where the re-projection error based solely on 2D human keypoints fails to capture the interaction between the human body and the scene, we introduce a 2D HVIP loss term, $E_{\text{hvip2D}}$. This loss term optimizes the spatial position of the human body by penalizing unrealistic placements of keypoints in relation to the scene, thereby better describing the interaction between the human body and the environment. The specific formula is as follows:
\begin{gather}
\label{eq:hvip_2d}
E_\text{hvip2D} = \lambda_{\text{h}}\sum_{n,t} \zeta _{n,t} \delta_{n,t} \left\| \Pi_\mathcal{K}(\mathcal{H} (\tilde{\mathbf{J}}_{n,t}^\text{3D},\bar{\mathbf{G}}))-\bar{\mathbf{p}}_{n,t}^\text{hvip2D}\right\|_2^2,
\end{gather}
where $\zeta_{n,t} \in \{0,1\}$ indicates whether the $n$-th individual in the $t$-th frame is valid. The function $\mathcal{H}(\cdot ,\bar{\mathbf{G}})$ is employed to compute the 3D human virtual interaction point on the ground plane $\bar{\mathbf{G}}$. This loss function primarily focuses on the vertical projection point between the human body's center of mass and the ground plane, but it neglects the actual distance between the human body and the ground. To address this, we introduce \( E_\text{contact} \) to further determine a reasonable position by minimizing the distance between the lowest point of the mesh and the ground. The specific definition is as follows:
\begin{gather}
\label{eq:trans_simplified}
E_\text{contact} = \lambda_{\text{c}}\sum_{n,t} \zeta _{n,t} \left\|\min\left\{\frac{|\bar{N}_\text{g}^T\tilde{\mathbf{V}}_{n,t}^{(i)} + \bar{D}_\text{g}|}{\|\bar{N}_\text{g}\|_2}, \forall i \right\}\right\|_1,
\end{gather}
where $\left\| \cdot \right\|_1$ denotes the $L1$ norm. $\tilde{\mathbf{V}}_{n,t}^{(i)}$ denotes the $i$-th vertex point of the mesh vertices of person $n$ at time $t$.
We also use a penalty term $E_\text{t-trans}$ to penalize its movement in the time domain $(\bigtriangleup_t)$, thereby better fitting the root translation $\tilde{\boldsymbol{\tau}}_{n,t}\in \mathbb{R}^3$. The specific definition is as follows:
\begin{gather}
\label{eq:trans}
E_\text{t-trans} = \lambda_{\text{t-trans}} \sum_{n,t} \zeta _{n,t}\left\|\bigtriangleup_t(\tilde{\boldsymbol{\tau}}_{n,t})\right\|_2^2,
\end{gather}
where $\bigtriangleup_t(\boldsymbol{\tau}_{n,t})=\boldsymbol{\tau}_{n,t}-\boldsymbol{\tau}_{n,t-1}$. This term allows us to obtain temporally more stable root translation with significantly less jitter.

The SMPL optimization obtains a plausible pose of human body by fitting the parameters of SMPL model.
\wh{In addition to the root translation $\boldsymbol{\tau}_{n,t}$ and root rotation $\boldsymbol{\gamma}_{n,t}$ variables, we also optimize the shape parameters $\boldsymbol{\beta}_{n,t}$ and pose parameters $\boldsymbol{\theta}_{n,t}$. Due to the inherent ambiguity in mapping 2D features to 3D poses, similar to \cite{pavlakos2019expressive, ye2023decoupling}, we employ the VPoser prior~\cite{SMPL-X:2019} to avoid generating impossible human poses. We optimize the latent variable $\boldsymbol{z}_{n,t}^{\varphi}$, which is obtained by transforming pose parameters through the VPoser encoder, and use its decoder to generate new pose parameters. }
The objective function for this stage is defined as:
\begin{gather}
  \label{eq:smpl_prior}
\begin{aligned}\begin{array}{c}\arg\min\\\boldsymbol{\tau}_{n,t}, \boldsymbol{\gamma}_{n,t} ,\boldsymbol{\beta}_{n,t},\boldsymbol{z}_{n,t}^{\varphi} : \forall n,t\end{array}E_{\text{smpl}}, \quad \text{with}\\
  E_{\text{smpl}} = E_{\text{root}}+ E_{\text{pose}} +  E_{\text{shape}} +  E_{\text{t-pose}}.
\end{aligned}
\end{gather}
The pose regularization term $E_\text{pose}$ computes the loss of the latent variable based on the prior assumption of a Gaussian distribution, serving as a penalty term:
\begin{gather}
\label{eq:vposer}
E_\text{pose} = -  \lambda_{{\varphi}}\sum_{n,t} \log \mathcal{N}(\tilde{\boldsymbol{z}}^{\varphi}_{n,t}; 0, I),
\end{gather}
where $\tilde{\boldsymbol{z}}^{\varphi}_{n,t}\in \mathbb{R}^{32}$ is the latent variable to be optimized. The shape regularization term $E_\text{shape}$ penalizes overly large shape parameters $\tilde{\boldsymbol{\beta}}_{n,t}$ to maintain the model's shape integrity and prevent the generation of unnatural shapes, which is defined as:
\begin{gather}
\label{eq:shape}
E_\text{shape} = \lambda_{\text{shape}}\sum_{n,t} \zeta _{n,t}\left\|\tilde{\boldsymbol{\beta}}_{n,t}\right\|_2^2.
\end{gather}
To minimize pose jitter in the time series and achieve smoother motion, we introduce a temporal pose penalty:
\begin{gather}
\label{eq:t-pose}
E_\text{t-pose} = \lambda_{\text{t-pose}}\sum_{n,t} \zeta _{n,t}\left\|\bigtriangleup_t(\tilde{\mathbf{J}}_{n,t}^\text{3D})\right\|_2^2.
\end{gather}
This term captures the essence of maintaining stability across temporal poses by penalizing excessive changes in the 3D joint configurations.


In the fundamental individual optimization process, we implement a two-stage optimization strategy to progressively enhance the accuracy of crowd motion in terms of position and posture. This involves applying various position-related constraints, such as 2D HVIP constraints, contact constraints, and 2D pose loss of the human, thereby providing reasonable initial values for subsequent optimization steps.

\subsection{Coarse-to-fine Group-guided Motion Optimization}
\label{sec:group-opt}
Our goal is to enhance the temporal stability of crowd motion and to recover poses that have been occluded for an extended period. To achieve this, we design a coarse-to-fine group-guided motion optimization strategy. This strategy initiates the optimization process with a \yl{Variational Autoencoder (VAE)}-based human motion prior and continues with a segment-level group-guided optimization, progressively reconstructing the motion of the crowd.

\textbf{Optimization with VAE-based Human Motion Prior.} 
To generate smooth and realistic human motion while addressing short-term occlusions, we integrate prior knowledge of human motion by developing a transformer-based VAE\cite{chen2023executing} as our motion prior model, denoted as $\mathcal{V}={\mathcal{E}_p, \mathcal{E}_r, \mathcal{D}}$. Unlike transition-based motion priors \cite{rempe2021humor}, our approach enables faster inference, which is crucial for handling large groups of people in extensive scenes.
We decompose human motion into local and global components, where local motion refers to body posture features, and global motion captures the root trajectory. 
We segment an individual's entire motion sequence into multiple segments, with each segment delineated as a distinct motion segment, each lasting 64 frames at a frame rate of 30 frames per second \yl{(so approximately 2 seconds)}.
For an individual $n$ in the $s$-th temporal segment, the local transformer encoder $\mathcal{E}_p$ and the global MLP encoder $\mathcal{E}_r$ generate the combined latent motion representation $\boldsymbol{z}_{n,s}^{\vartheta} = (\boldsymbol{z}_{n,s}^p, \boldsymbol{z}_{n,s}^r)$, where $\boldsymbol{z}_{n,s}^{\vartheta} \in \mathbb{R}^{256 + 128}$. These representations serve as optimization variables in the coarse-to-fine group-guided motion optimization, allowing the decoder $\mathcal{D}$ to effectively recover motion sequences.
The state of a moving person is represented as $\boldsymbol{x}=(\boldsymbol{x}_p, \boldsymbol{x}_r)$, $\boldsymbol{x}\in\mathbb{R}^{330}$ at time $t$. The local motion state $\boldsymbol{x}_p=(\boldsymbol{\hat{J}},\boldsymbol{\hat{{\nu}}}, \boldsymbol{\hat{\theta}})$  \yl{is composed} of joint positions $\boldsymbol{\hat{J}} \in \mathbb{R}^{21\times3}$, joints velocity $\boldsymbol{\hat{{\nu}}} \in \mathbb{R}^{21\times3}$ and joint rotations $\boldsymbol{\hat{\theta}}\in \mathbb{R}^{21\times9}$, where the joints are in local coordinate system 
that the local joints are relative to the current frame pelvis joint and retains the position coordinates of the ground direction.
The global motion state $\boldsymbol{x}_r=(\boldsymbol{\hat{r}},\boldsymbol{\hat{\tau}},\boldsymbol{\hat{\gamma}})$ includes root positions $\boldsymbol{\hat{r}}\in\mathbb{R}^{3}$, the global translation $\boldsymbol{\hat{\tau}}\in\mathbb{R}^{3}$ and rotation $\boldsymbol{\hat{\gamma}}\in\mathbb{R}^{9}$ of SMPL~\cite{SMPL:2015}.
An additional output is the contact probability $c$, which represents the likelihood of each joint being in contact with the ground.
The contacts 
enable ground constraints in optimization, promoting the 
\yl{coherence}
of interaction with the ground. 
Consistent with prior approaches~\cite{rempe2020contact, zhang2021learning, guo2022generating, zhang2024rohm}, we employ KL divergence, and L2 loss of each state to achieve training from AMASS dataset~\cite{mahmood2019amass}.
To enhance the occlusion completion ability of our prior model to the occlusion 
\yl{commonly}
existing in large scenes, we define the visibility masks $\boldsymbol{m_p}\in\{0,1\}^{330}$ of input motion state.
During the training phase, we \yl{synthesize} various occlusion masks on \yl{the} AMASS dataset, including random joint occlusion and frame occlusion, partial body occlusion, and random consecutive occlusion for global or local input states.
In the optimization phase, we set the masks \yl{based on} the confidence of the detected 2D joints.


We apply our motion prior model to optimization to obtain reasonable motion trajectories and postures.
The optimization variables consist of the SMPL shape parameters for each segment of the motion sequence, denoted as $\boldsymbol{\beta}_{n,s}$, and the latent variables $\boldsymbol{z}_{n,s}^{\vartheta}$. The objective function for this stage is defined as:
\begin{gather}
  \label{eq:motion_prior}
\begin{aligned}&\begin{array}{c}\arg\min\\\boldsymbol{\beta}_{n,s},\boldsymbol{z}_{n,s}^{\vartheta} : \forall n,s\end{array}E_{\text{motion}}, \quad \text{with}\\
  E_{\text{motion}} &= E_{\text{smpl}}+ E_{\text{VAE}} +  E_{\text{env}} + E_{\text{connect}}.
\end{aligned}
\end{gather}
In order to make the motion reasonable, $E_\text{VAE}$ optimizes the latent variable based on the prior assumption of the Gaussian distribution as a penalty term:
\begin{gather}
\label{eq:vae}
E_\text{VAE} = -  \lambda_{\text{vae}}\sum_{n,s} \log \mathcal{N}(\tilde{\boldsymbol{z}}_{n,s}^{\vartheta}; 0, I),
\end{gather}
where $\tilde{\boldsymbol{z}}_{n,s}^{\vartheta}$ represents the latent variables to be optimized. The environmental energy $E_\text{env}$ ensures that each individual maintains a specified height range from the ground upon contact, with a velocity of zero at the moment of contact. This is expressed as: 
\begin{gather}
\label{eq:hight}
\begin{aligned}
E_\text{env} = \sum_{n,t_s,j}&\lambda_{\text{ch}}c_{n,t_s,j}\max(0, |\tilde{\mathbf{J}}_{n,t_s,j}^{\text{3D},z} - h|)+\\
&\lambda_{\text{cv}}c _{n,t_s,j}\left\|\bigtriangleup_t(\tilde{\mathbf{J}}_{n,t_s,j}^\text{3D})\right\|_2^2,
\end{aligned}
\end{gather}
where $t_s$ is the time frame of the $s$\yl{-th} segment, $c_{n,t_s,j}$ is the contact probability of the $j$-th joint of the $n$-th individual in the $t_s$-th frame of the model output. The contact height term consists of the $z$-coordinate component of the contacting joint being no more than $h$ above the ground in the standard frame. $E_\text{connect}$ serves as a penalty term to ensure motion consistency between the end of one segment and the beginning of the adjacent segment within the same sequence:
%
\begin{gather}
\label{eq:connect}
\begin{aligned}
E_\text{connect} =& \lambda_\text{con} \sum_{n,s} \zeta_{n,s_{l}} \zeta_{n,s^+_1} \left\|\tilde{\mathbf{J}}_{n,s^+_1}^\text{3D} - \tilde{\mathbf{J}}_{n,s_{l}}^\text{3D}\right\|_2^2,
\end{aligned}
\end{gather}
where $s_{l}$ is the last frame of the current segment, and $s^+_1$ is the first frame of the next segment. This constraint term ensures continuous motion from segment to segment.

\begin{figure}[t!]
\centering
\includegraphics[width=\linewidth]{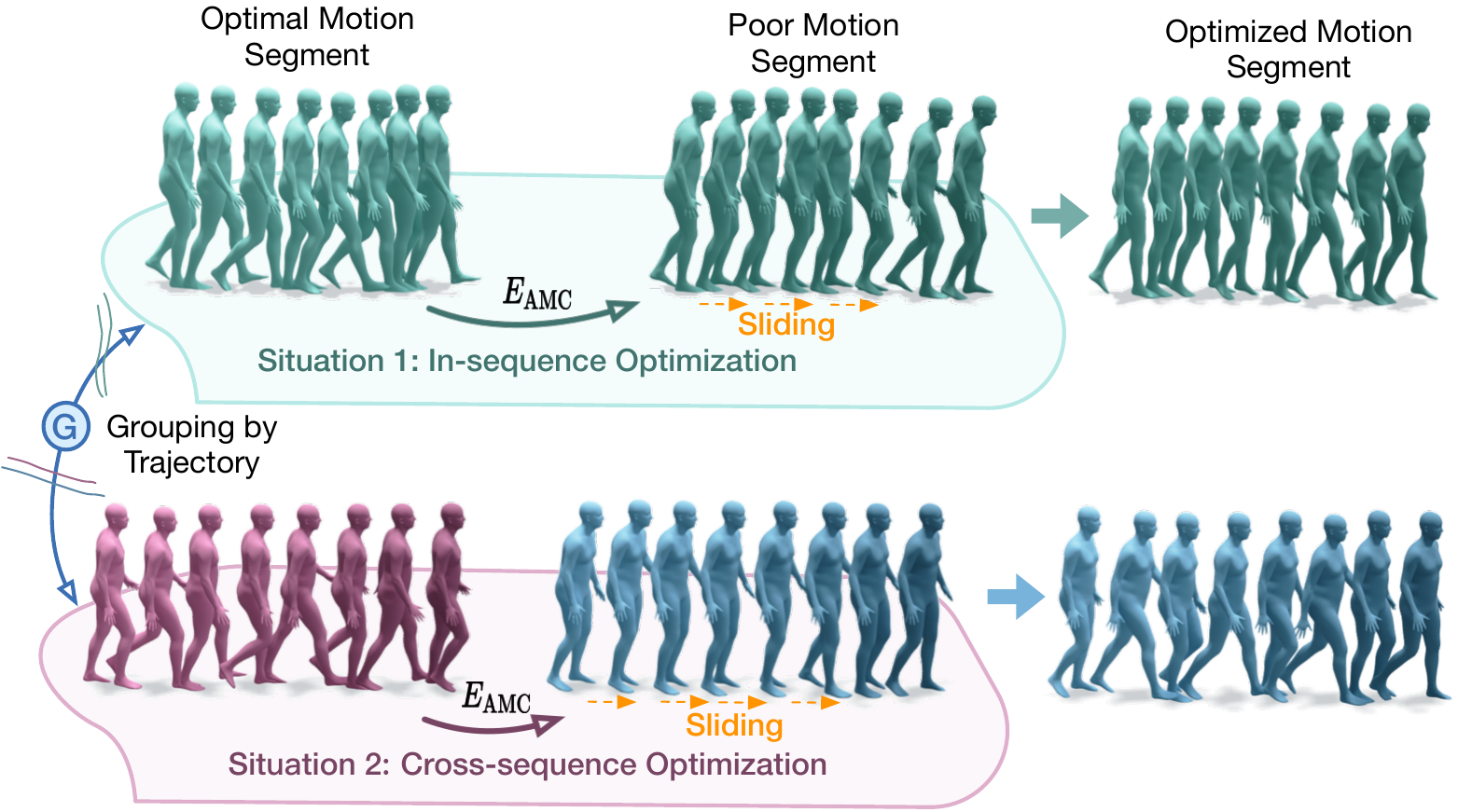}
\caption{Segment-level group-guided optimization. Motion segments are grouped based on similar relative trajectories, with the optimal motion segments in each group assisting in the optimization of heavily occluded segments. It includes both in-sequence and cross-sequence optimization situations, depending on whether the optimal motion sequence belongs to the same individual.}
\label{fig:fig3}
\end{figure}

\textbf{Segment-level Group-guided Optimization.}
The motion prior model helps obtain temporally stable human motion and can recover from short-term and sporadic occlusions, but struggles to handle severe long-term occlusions. To address this limitation, we propose a segment-level group-guided optimization method that enhances the motion recovery of long-term occluded sequences by leveraging the collectivity of crowd behavior.
\khb{As shown in Fig. \ref{fig:fig3}, unlike the GroupRec~\cite{huang2023reconstructing} method, which groups based on collective behavior in static images, we first group the segments according to similar relative motion trajectories. Following this, we jointly optimize their motion sequences using our Asynchronous Motion Consistency (AMC) loss (see below).}
Within each group, the unoccluded segments are designated as the optimal motion segments, serving to guide those with poor motion segments.
Based on whether the guiding and guided motion segments belong to the same individual, we can categorize the scenarios into two situations: in-sequence and cross-sequence. 
In the in-sequence scenario, the sequence containing occlusions is recovered through guidance from its own unobstructed segments.

\wh{During the grouping process, the symmetric segment path distance~\cite{besse2015review} is adopted as the distance metric between motion trajectories. This approach allows us to focus more on the overall shape of the trajectories while accommodating variations in trajectory lengths. For clustering, we employ the affinity propagation clustering~\cite{frey2007clustering} algorithm, which eliminates the need to predefine the number of clusters.}

To achieve the guidance optimization between motion segments within a group, we design an asynchronous motion consistency loss to quantify the differences in intra-group motion.
It includes a motion segment confidence to identify the optimal and poor segments within a group. 
We calculate the motion segment confidence $c_{n,s}^{\text{seg}}$ for each motion segment in a group based on detection scores and joint importance:
\begin{gather}
\label{eq:seg_conf}
\begin{aligned}
c_{n,s}^\text{seg}={v}_{n,s}
\cdot \left(1 - \mathds{1}(\sum_{j} w_j \mathcal{F}_j(\bar{\mathbf{J}}_{n,s}^\text{2D}) > a)\right),
\end{aligned}
\end{gather}
\wh{where $v_{n,s}$ represents the proportion of visible joints (those with high detection scores) in the motion segment, $w_j$ denotes the predefined joint weight based on the importance of joint occlusion, and $\mathcal{F}_j(\cdot)$ is a decision function used to determine whether the $j$-th joint is continuously occluded.}
The parameter $a$ is a threshold that signifies the need for repair. Based on the scores of motion segment confidence, we adaptively determine the segments that require repair and identify high-quality segments in a group:
\begin{gather}
\label{eq:xi}
\begin{aligned}
\xi_{n,s} = 
\begin{cases} 
1 & \text{if } c_{n,s}^\text{seg} = \max({\{c_{n,s}^\text{seg}\}}_{s=1}^{S_g}) \\
0 & \text{if } c_{n,s}^\text{seg} = 0 \\
-1 & \text{otherwise}
\end{cases}\quad,
\end{aligned}
\end{gather}
where $S_g$ is the number of segments within the group. When the value of $\xi_{n,s}$ is 1, it indicates the optimal motion segment; a value of 0 denotes a poor motion segment; and -1 signifies a segment that does not require optimization. 
The collective nature of intra-group motion is utilized for motion recovery, using unobstructed segments (i.e., those where $\xi_{n,s} = 1$) to restore the motion of severely obscured segments (i.e., those where $\xi_{n,s} = 0$).
If there exists more than one $\xi_{n,s}=1$, we prioritize using the optimal segment in the in-sequence scenario.

While the motions within a group appear similar in general, they often exhibit temporal misalignments and rhythmic variations in their posture sequences. 
Innovatively, we treat each individual's motion sequence as a time series and propose our asynchronous motion consistency loss $E_{\mathrm{AMC}}$ as follows:
\begin{gather}
\label{eq:amc}
\begin{aligned}
E_{\mathrm{AMC}}=\lambda_\text{AMC}\sum_{n,s}&w\mathcal{L}_\text{s}(\tilde{\boldsymbol{\theta}}_{n,s},\tilde{\boldsymbol{\theta}}_{n',s'})\cdot\mathds{1}({\xi}_{n,s}=0,{\xi}_{n',s'}=1).
\end{aligned}
\end{gather}
\wh{Here, $\tilde{\boldsymbol{\theta}}_{n,s}$ and $\tilde{\boldsymbol{\theta}}_{n',s'}$ represent the pose parameters of the poor segment and the optimal segment, respectively. The term $w$ denotes the predefined joint weights. To preserve the flexibility of individual motions, we exclude joints associated with expressive details, such as those in the head and hands.
The function $\mathcal{L}_\text{s}$ corresponds to the Soft Dynamic Time Warping (sDTW)~\cite{cuturi2017soft} method, a differentiable variant of Dynamic Time Warping (DTW)~\cite{sakoe1978dynamic} . DTW is a dynamic programming algorithm designed to measure the similarity between two time series through nonlinear alignment, enabling it to handle sequences with differing speeds or lengths. In our approach, we employ sDTW to temporally stretch and align the joint rotations of motion segments within the group. }
The proposed asynchronous motion consistency loss measures the variance in motion within a group, facilitating the transition from occluded poses to high-quality poses, which is beneficial for motion recovery during periods of long-term occlusion.

We set $\boldsymbol{\beta}_{n,s}$, and $\boldsymbol{z}_{n,s}^{\vartheta}$ as optimization variables,  and the objective function for optimization is:
\begin{gather}
  \label{eq:motion_prior2}
\begin{aligned}\begin{array}{c}\arg\min\\\boldsymbol{\beta}_{n,s},\boldsymbol{z}_{n,s}^{\vartheta} : \forall n,s\end{array}E_{\text{group}}, \quad \text{with}\\
  E_{\text{group}} = E_{\text{motion}}+ E_{\text{AMC}}.
\end{aligned}
\end{gather}


In our coarse-to-fine group-guided motion optimization process, we first leverage prior knowledge of human motion to address the issue of instability in human motion. Subsequently, we employ a segment-level group-guided optimization with asynchronous motion consistency loss to tackle the problem of pose loss due to long-term occlusions, progressively reconstructing natural and complete human motion.
As a result, the optimized individuals, featuring spatio-temporally consistent locations and accurate human poses, constitute the reconstruction output of the DyCrowd framework.



\section{Experiments}

\subsection{Datasets and Evaluation Metrics}
\label{sec:datasets}

\begin{figure*}[t!]
\centering
\includegraphics[width=\textwidth]{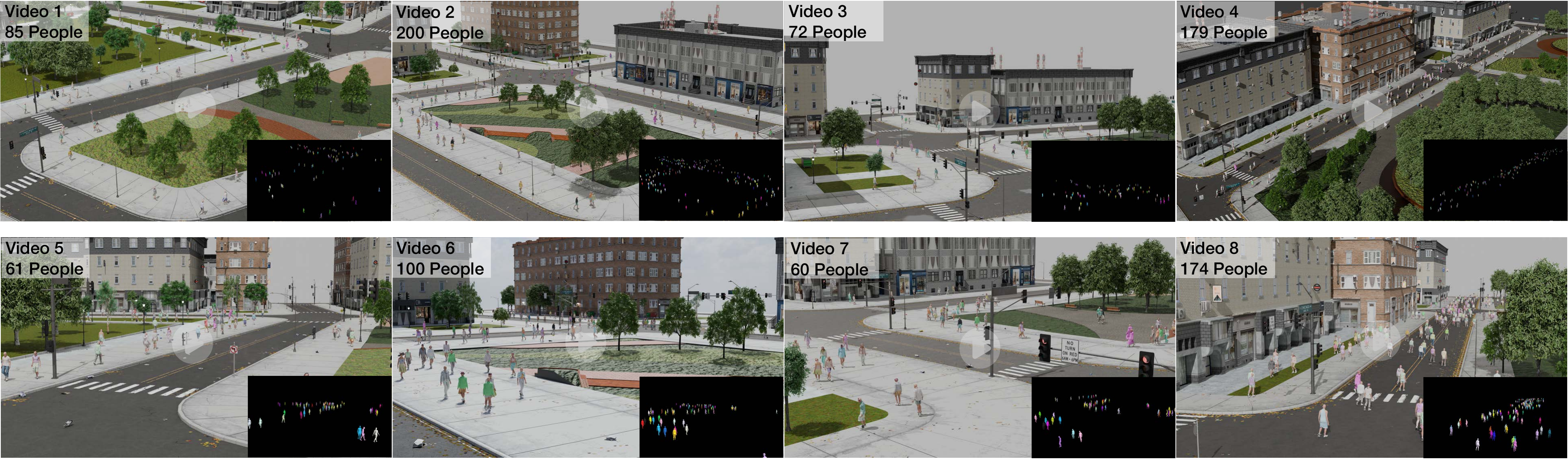}
\caption{\khb{Eight validation videos in the \textit{VirtualCrowd} dataset and the corresponding crowd distributions using the SMPL-X model. The first row is the high-angle view, and the second row is the low-angle view.}}
\label{fig:dataset}
\end{figure*}


\textbf{Virtual Large-scene Dataset.}  
To validate our proposed method, we create a comprehensive large-scale dynamic crowd dataset named \textit{VirtualCrowd}. This dataset utilizes Blender's ICity plugin \cite{icity3d} to construct diverse scenes and extracts ground plane information to generate human motion trajectories. By employing SynBody \cite{yang2023synbody} human models and the DIMOS \cite{zhao2023synthesizing} algorithm, we generate a variety of motion sequences. Subsequently, we render the dataset using Blender, resulting in a large-scale crowd motion video dataset, as shown in Fig. \ref{fig:dataset}. 
\khb{The dataset consists of four scenes, each covering an area larger than 2500 square meters. For each scene, two configurations are provided, incorporating both high-angle and low-angle perspectives, with varying numbers of people in different configurations. The high-angle perspective offers a broader field of view, while the low-angle perspective has more severe occlusions and greater scale differences between individuals. The dataset ultimately includes eight validation videos, each with a resolution of 7680×4320 (8K) and a frame rate of 30 frames per second, totaling 1600 frames. The scenes exhibit varying crowd densities, with the number of people ranging from 60 to 200. In total, there are 931 motion sequences and 186,200 poses.}
Furthermore, the dataset is equipped with detailed 2D and 3D annotations, including 2D joints, MOT annotations, 3D joints, 3D positions, and SMPL-X parameters, for evaluating dynamic crowd reconstruction in large scenes.  
\textit{VirtualCrowd} provides essential foundational data support for crowd reconstruction, behavior analysis, and related research.

\kkhb{Furthermore, to further enrich the dataset and increase the diversity of scenes, we have extended the dataset with two videos of a sloped scene featuring different perspectives and crowd distributions, based on simulation techniques from \cite{liu2025rescue}, thus providing support for subsequent research. For more details, please refer to the supplementary material.}

\textbf{Real Large-scene Dataset.}  
PANDA~\cite{wang2020panda} is a gigapixel-level human-centric video dataset captured by a gigapixel camera and covering real-world scenes. As PANDA does not contain the poses of persons, we conduct qualitative comparisons on the PANDA dataset.

\textbf{Evaluation Metrics.} 
To verify the consistency of the reconstructed crowd in global space, we follow \cite{wen2023crowd3d} and use the Procrustes-aligned pair-wise percentual distance similarity (PA-PPDS) to evaluate the location distribution of the crowd based on the distances among people.
We also use the percentage of correct ordinal depth (PCOD)~\cite{zhen2020smap} to measure the correctness of ordinal depth relation.
To evaluate 3D pose estimation performance, we employ mean per joint position error (MPJPE) and Procrustes-aligned MPJPE (PA-MPJPE). We focus on the accuracy of pose sequences in the global space and report two metrics: WA-MPJPE and W-MPJPE. These metrics~\cite{ye2023decoupling} align each output sequence with the ground-truth data using the entire sequence and the first two frames, respectively.
To measure the motion smoothness, we compute acceleration error (ACCEL) against the ground-truth acceleration.

\begin{table*}[!t]
    \caption{\khb{Quantitative Comparison on VirtualCrowd Dataset.}}
    \centering
    \small
    \begin{threeparttable} 
    \begin{tabular}{@{}c|ccccccc@{}}
    \toprule
    Method   &PA-PPDS$\uparrow$    & PCOD$\uparrow$      &MPJPE$\downarrow$   &PA-MPJPE$\downarrow$  &WA-MPJPE$\downarrow$  &W-MPJPE$\downarrow$ &ACCEL$\downarrow$ \\
    \midrule
    Crowd3D \cite{wen2023crowd3d}    & 87.98                 & 91.11       & 122.99   &73.70     & -                 & -       & -   \\
    GroupRec \cite{huang2023reconstructing}       & 75.04      & 86.22         & 89.04     &58.98    & 82.20               & 94.34      & 165.50     \\
    DyCrowd      & \textbf{89.10}        & \textbf{92.20}       & \textbf{69.74}     &\textbf{48.57}     & \textbf{68.99}                 & \textbf{83.39}      & \textbf{15.72}  \\
    \hline
    $\text{SLAHMR-Large}^{\star}$\cite{ye2023decoupling}    & 87.95      & 90.34        & 106.35     & 69.41    & 90.10                 & 108.40       & \textbf{12.25}      \\
    
    $\text{DyCrowd}^{\star}$       & \textbf{91.23}         & \textbf{95.38}       & \textbf{68.81}      &\textbf{45.34}     & \textbf{65.91}                 & \textbf{80.34}       & 15.53    \\
    \bottomrule
    \end{tabular}
    \begin{tablenotes} 
    \item The symbol $\star$ denotes the use of ground-truth object tracking, while "-" means unavailable results. SLAHMR-Large is a variant of SLAHMR modified for large scenes, and we compare it in the ground-truth object tracking setting.
    
     \end{tablenotes} 
    \end{threeparttable} 
\label{tab_sota}
\end{table*}

\begin{table*}[t]
    \caption{\khb{Ablation Study Results on VirtualCrowd Dataset.}}
    \centering
    \small
    \begin{threeparttable} 
    \begin{tabular}{@{}c|ccccccc@{}}
    \toprule
    Method   &PA-PPDS$\uparrow$    & PCOD$\uparrow$      &MPJPE$\downarrow$   &PA-MPJPE$\downarrow$  &WA-MPJPE$\downarrow$  &W-MPJPE$\downarrow$ &ACCEL$\downarrow$ \\
    \midrule
    
    DyCrowd w/o C.G.M.            & \textbf{89.15}       & 92.18          & 86.87   &61.27     & 80.23                 & 88.57       & 73.33       \\
    DyCrowd w/o AMC                 & 89.12      & 92.19          & 70.22     &49.02     & 69.41                 & \textbf{83.35}      & \textbf{15.08}    \\
    DyCrowd      & 89.10        & \textbf{92.20}       & \textbf{69.74}     &\textbf{48.57}     & \textbf{68.99}                 & 83.39      & 15.72  \\
    \bottomrule
    \end{tabular}
    \end{threeparttable} 
\label{tab_stage}
\end{table*}

\begin{table}[t]
    \caption{\khb{Ablation Study Results for Occluded Instances in the VirtualCrowd Dataset.}}
    \centering
    \small
    \begin{tabular}{@{}c|cc@{}}
    \toprule
    Method   &MPJPE$\downarrow$   &PA-MPJPE$\downarrow$ \\
    \midrule
    
    DyCrowd w/o C.G.M.               & 73.01     & 60.20        \\
    DyCrowd w/o AMC                 & 67.17     & 56.77            \\
    DyCrowd      & \textbf{66.09}       & \textbf{56.21}       \\
    \bottomrule
    \end{tabular}
\label{tab_occlusion}
\end{table}

\begin{figure*}[t!]
\centering
\includegraphics[width=\textwidth]{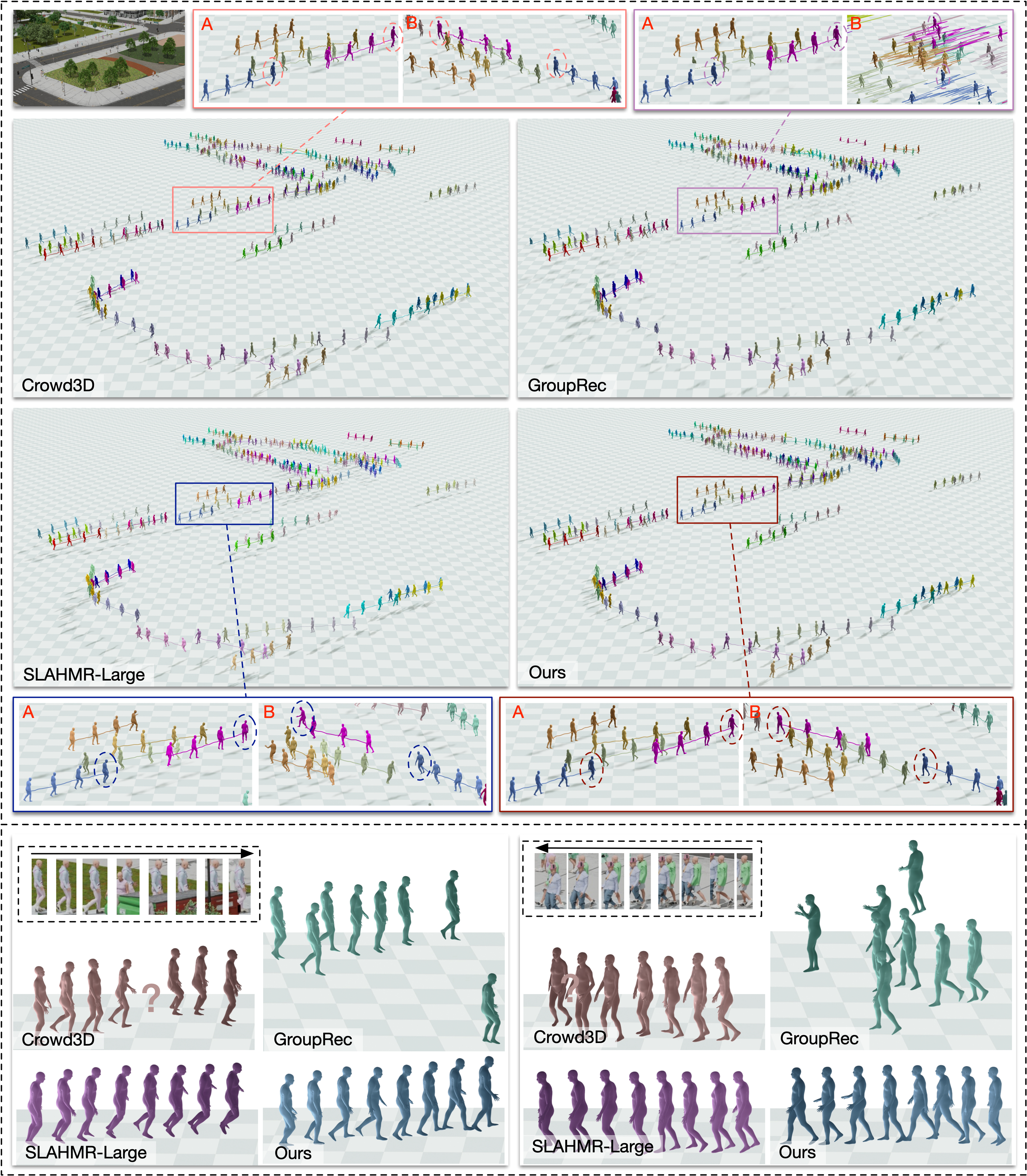}
\caption{ Qualitative comparison on the \textit{VirtualCrowd} dataset. The highlighted areas, including the camera perspective (A) and side perspective (B), demonstrate that our method achieves spatio-temporally consistent crowd reconstruction with plausible motion. Individual instance images demonstrate that our method is capable of restoring human motion even under severe occlusions.}
\label{fig:virtual}
\end{figure*}

\begin{figure*}[t!]
\centering
\includegraphics[width=\textwidth]{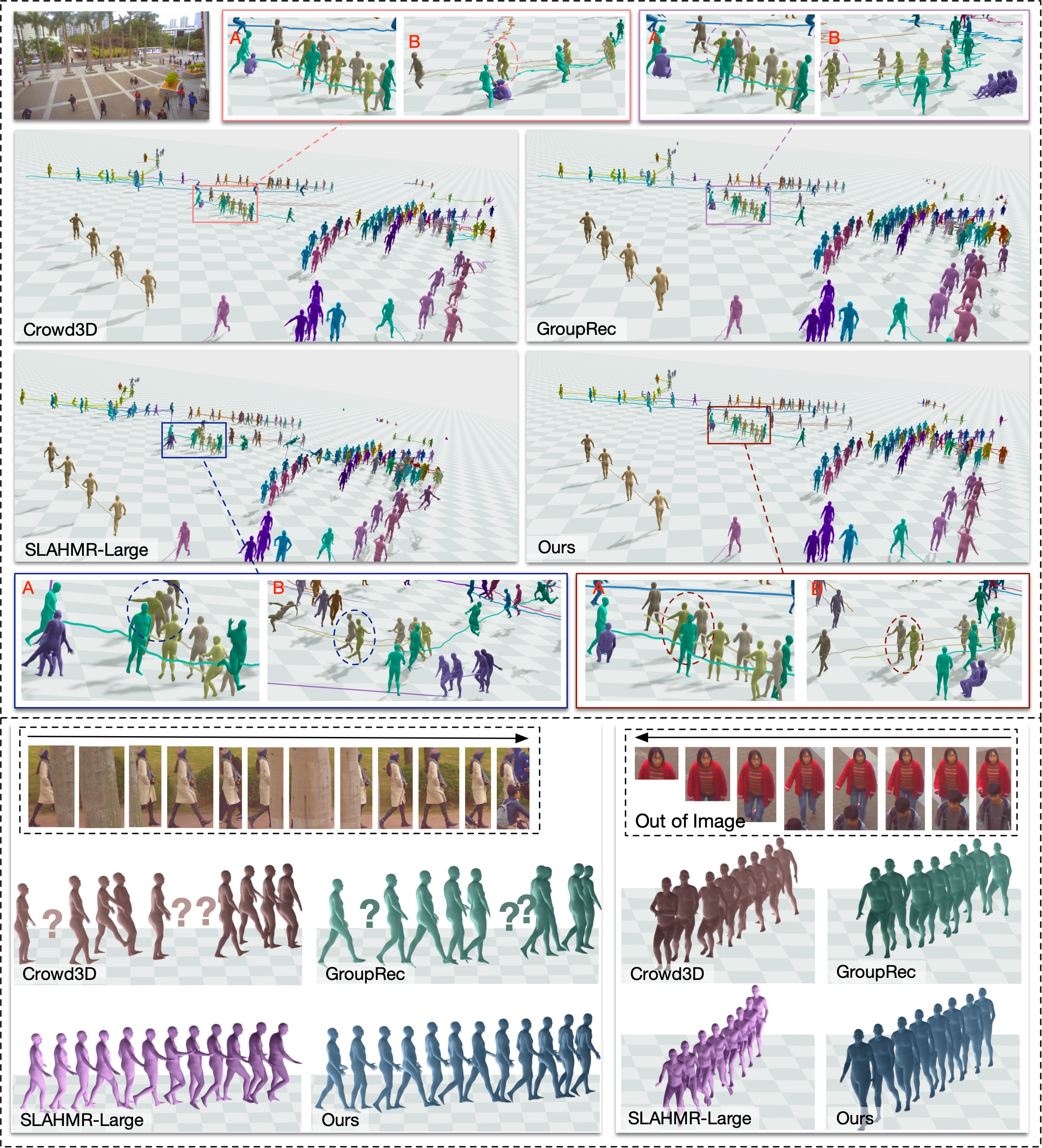}
\caption{Qualitative comparison on the \textit{PANDA} dataset.}
\label{fig:panda}
\end{figure*}

\subsection{Implementation Details}
We execute experiments using PyTorch~\cite{paszke2019pytorch}, employing the RMSprop~\cite{hinton2012neural} optimizer with a learning rate 0.01. We progressively carry out root optimization, SMPL optimization, optimization with a VAE-based human motion prior, and segment-level group-guided optimization, iterating 100, 150, 200, and 200 times respectively. The weights $\lambda$ used in the experiments are empirically defined to balance the magnitude of individual energy terms and remain constant throughout all experiments. 
We use $\lambda_{\text{2D}}$ = 100, 
$\lambda_{\text{h}}$ = 10000, 
$\lambda_{\text{c}}$ = 50, 
$\lambda_{\text{t-trans}}$ = 50, 
$\lambda_{{\varphi}}$ = 0.5, 
$\lambda_{\text{shape}}$ = 1, 
$\lambda_{\text{t-pose}}$ = 10, 
$\lambda_{\text{vae}}$ = 0.2, 
$\lambda_{\text{ch}}$ = 500, 
$\lambda_{\text{cv}}$ = 1000, 
$\lambda_\text{con}$ = 100, 
$\lambda_\text{AMC}$ = 0.03.
Optimizing a large-scene video with 100 people and 200 frames takes approximately 4 hours on a workstation equipped with an NVIDIA 3090 GPU and 128GB memory.

\subsection{Comparison}
Currently available methods are not designed for spatio-temporally consistent reconstruction of crowd motion in large-scene videos. Thus, we compare our approach with two state-of-the-art image-based large-scene methods: Crowd3D~\cite{wen2023crowd3d} and GroupRec~\cite{huang2023reconstructing}. Additionally, we modify the state-of-the-art video-based method SLAHMR~\cite{ye2023decoupling}, originally designed for multi-person reconstruction in small scenes, to extend its capabilities to large scenes, dubbing it SLAHMR-Large.
For fair comparison, we provide the scene-level camera focal length and ground plane parameters estimated by our method to them.
For Crowd3D and GroupRec, we process each image in the large-scene videos separately. To ensure consistency of the human sequence during evaluation, we assign the results of our object tracking to GroupRec, replacing its original detection input. 
Crowd3D does not participate in sequence-level evaluation because it lacks effective temporal matching between multiple targets across frames.
We modify SLAHMR by replacing its initialization with our global crowd motion initialization and using camera extrinsic parameters derived from our estimated ground plane. We perform the optimization of individuals in the scene in batches, due to the large scene and the high video memory requirements of SLAHMR. 
Since SLAHMR exhibits low tolerance for data noise, its optimization process may be disrupted when dealing with sequences that have poor detections or tracking. Additionally, due to the high computational cost—requiring approximately 8 hours to process every 25 people—we restrict our evaluations to scenarios with ground-truth object tracking.

As demonstrated in Table \ref{tab_sota}, our method outperforms existing methods across all evaluation metrics except for ACCEL. Our method achieves the best PA-PPDS and PCOD scores, indicating a superior spatial arrangement.
Moreover, benefiting from our coarse-to-fine group-guided motion optimization strategy, our method substantially surpasses existing methods in terms of pose estimation accuracy.
The small ACCEL value obtained by our method suggests smooth and coherent human motion. We observe that SLAHMR-Large yields even smoother predictions because it utilizes the motion prior HuMoR\cite{rempe2021humor}, which favors smooth transitions between consecutive frames. However, when processing long or occluded sequences, it often generates nonsensical sliding rather than accurate pose predictions.
Fig. \ref{fig:virtual} and Fig. \ref{fig:panda} show the qualitative comparison on \textit{VirtualCrowd} and \textit{PANDA}~\cite{wang2020panda} datasets. Our method achieves spatio-temporal consistency in the reconstruction of crowd motion, with reconstructed individuals exhibiting more natural and plausible postures. Particularly, by leveraging the collective \yl{behavior} of crowds in large scenes, our method is the only one capable of accurately restoring human motion in instances of long-term occlusion.

\subsection{Ablation Study}
We conduct ablation studies to investigate the effects of key designed elements. For ease of reference, we abbreviate DyCrowd without Coarse-to-fine Group-guided Motion optimization to DyCrowd w/o C.G.M. and DyCrowd without Asynchronous Motion Consistency loss to DyCrowd w/o AMC.



\khb{\textbf{Coarse-to-fine Group-guided Motion Optimization.}
By comparing DyCrowd w/o C.G.M. and DyCrowd without AMC in Table \ref{tab_stage}, it can be seen that our human motion prior model yields more accurate pose estimations. Since the AMC loss mainly focuses on dynamic occlusion recovery and local sequence consistency, it may cause slight fluctuations in global metrics such as PA-PPDS and W-MPJPE, and the introduction of more realistic but not completely smooth motion details leads to a slight increase in ACCEL. We further present the pose metrics for severely occluded instances in Table \ref{tab_occlusion}, which demonstrate that our method achieves the best performance under severe occlusion, significantly improving robustness to occlusion.}

\begin{figure}[t]
\centering
\includegraphics[width=.9\linewidth]{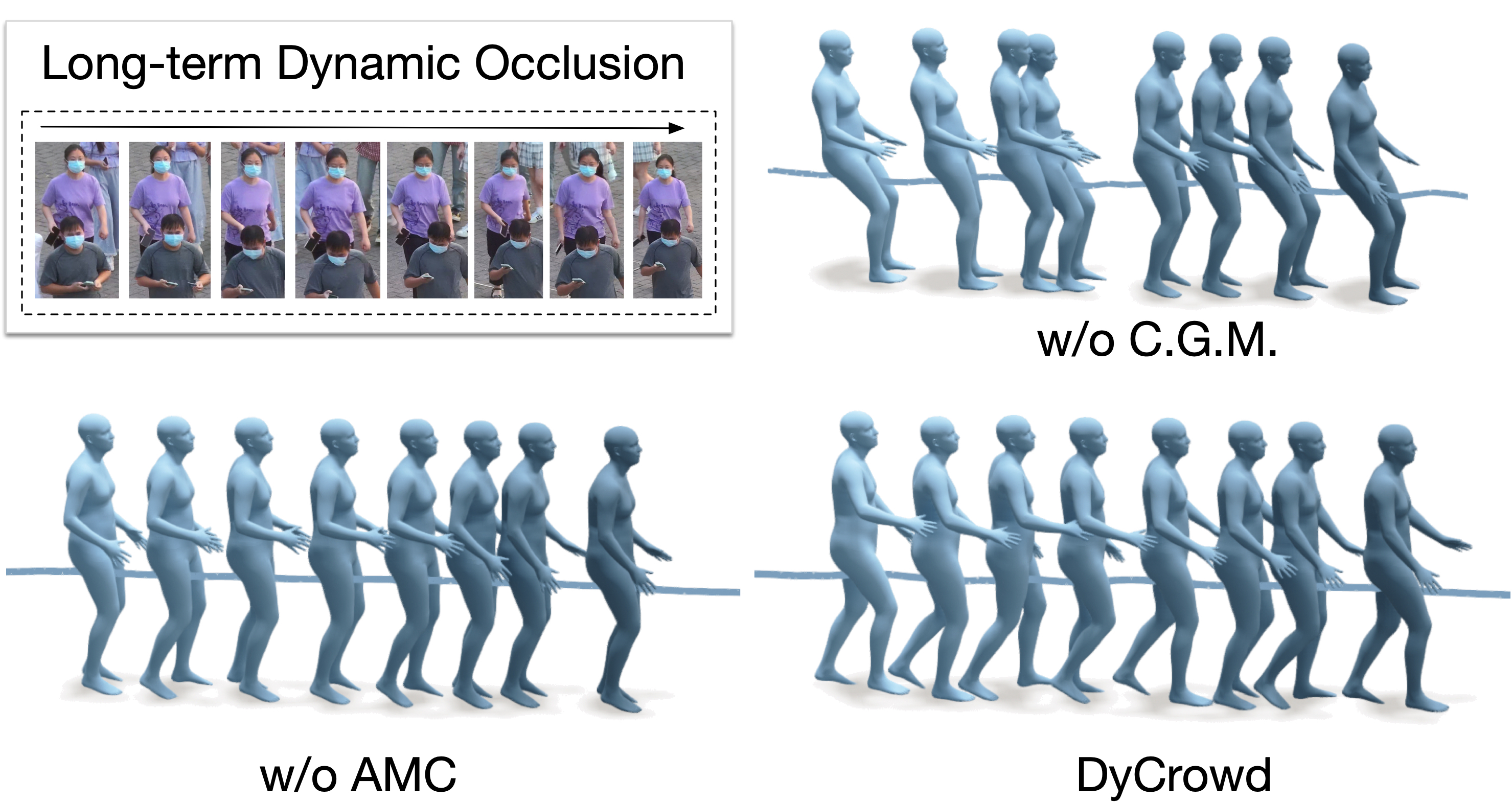}
\caption{Qualitative comparison of our key designs for occlusion.}
\label{fig:stage}
\end{figure}

\textbf{Asynchronous Motion Consistency Loss.} 
Our asynchronous motion consistency loss enables high-quality unoccluded motion segments to guide the motion recovery of occluded ones within groups that exhibit similar motion trajectories. This approach ensures robust and plausible motion recovery even under severe occlusion. Although the results in Table \ref{tab_stage} indicate a slight compromise in performance compared to DyCrowd w/o AMC, Fig. \ref{fig:stage} demonstrates that the motion under long-term occlusion is effectively restored by AMC loss, which we consider more meaningful. Additionally, Table \ref{tab_occlusion} \yl{showcases} that AMC loss further enhances the robustness against occlusion.

\begin{table*}[!t]
    \caption{\khb{Ablation experiments on different VAE-based motion priors in the VirtualCrowd dataset.}}
    \centering
    \small
    \begin{tabular}{@{}c|ccccccc@{}}
    \toprule
    Method   &PA-PPDS$\uparrow$    & PCOD$\uparrow$      &MPJPE$\downarrow$   &PA-MPJPE$\downarrow$  &WA-MPJPE$\downarrow$  &W-MPJPE$\downarrow$ &ACCEL$\downarrow$ \\
    \midrule
    w/ NeMF~\cite{he2022nemf}            & 88.72       & 91.86          & 84.63   &59.43     & 80.64                 & 99.44       & \textbf{13.87}       \\
    w/ DMMR-VAE~\cite{huang2023simultaneously}                 & 89.09      & 92.16          & 84.13     &58.71     & 78.14                 & 87.08     &157.69    \\
   Ours      & \textbf{89.12}      & \textbf{92.19}          & \textbf{70.22}     &\textbf{49.02}     & \textbf{69.41}                 & \textbf{83.35}      & 15.08    \\
    \bottomrule
    \end{tabular}
\label{other_vae}
\end{table*}

\khb{\textbf{VAE-based Motion Prior.} 
To validate the effectiveness of the VAE, we compared different VAE architectures. Specifically, we chose NeMF~\cite{he2022nemf} and DMMR-VAE~\cite{huang2023simultaneously} as baselines, as these methods have recently been shown to be effective in human motion optimization tasks~\cite{huang2023simultaneously,kocabas2024pace}. }
\khb{As shown in Table \ref{other_vae}, our VAE exhibits a significant advantage in pose reconstruction accuracy. These results stem from our key design for crowd motion optimization in large-scale scenes. Although the ACCEL index is slightly higher than that of some methods in certain cases, it mainly reflects the smoothness of motion. However, our method emphasizes retaining the real dynamic motion in complex large-scale scenes to obtain more motion details.}


\section{CONCLUSION AND DISCUSSION}
\textbf{Conclusion.}
We propose DyCrowd, the first framework to reconstruct spatio-temporally consistent 3D positions, poses, and shapes of crowd in a unified global space from a large-scene video. Our method incorporates prior knowledge of human motion and leverages the collective behavior of crowds in large scenes to facilitate the optimization of occluded human motion guided by visible human motion. This enables our method to not only reconstruct natural and continuous human motion but also achieve recovery of motion under long-term occlusion.
We also contribute a virtual large-scene evaluation dataset named \textit{VirtualCrowd} to facilitate future research on dynamic crowd reconstruction and analysis in large scenes.

\textbf{Limitations and Future Work.}
Although our method achieves the reconstruction of crowd motion in large-scene videos, there are still some limitations. Firstly, our method may be severely affected in cases of catastrophic failures in 2D keypoint detection and tracking. For instance, when a tracklet of a person is interrupted due to occlusion, we cannot recover the occluded individual's motion. Secondly, although our method can handle multi-ground scenes by region-level processing, it falls short in accurately capturing human interactions with complicated ground planes, such as climbing stairs. Lastly, our method employs a multi-stage optimization framework, which does not currently support real-time applications.
In the future, we plan to integrate scene and environmental information into our framework. This will advance our ability to recover human motion in complex scene interactions.

\section*{Acknowledgments}
This work was supported in part by National Key R\&D Program of China (2023YFC3082100), National Natural Science Foundation of China (62171317), and Science Fund for Distinguished Young Scholars of Tianjin (No. 22JCJQJC00040).



 
%
\normalem
\bibliography{egbib}
\bibliographystyle{IEEEtran}

\vfill

\end{document}